\documentclass{article}

\usepackage[usenames,dvipsnames,svgnames,x11names]{xcolor}

\usepackage[nocompress]{cite}

\usepackage{listings}

\lstdefinelanguage{XML}
{
basicstyle=\ttfamily\footnotesize,
  morestring=[b]",
  moredelim=[s][\bfseries\color{Maroon}]{<}{\ },
  moredelim=[s][\bfseries\color{Maroon}]{</}{>},
  moredelim=[l][\bfseries\color{Maroon}]{/>},
  moredelim=[l][\bfseries\color{Maroon}]{>},
  morecomment=[s]{<?}{?>},
  morecomment=[s]{<!--}{-->},
  commentstyle=\color{gray},
  stringstyle=\color{blue},
  identifierstyle=\color{red}
}
\usepackage{moreverb}
\usepackage{makecell}

\usepackage[nounderscore]{syntax}

\usepackage[pdftex]{graphicx}
\graphicspath{{./figures/}}
\DeclareGraphicsExtensions{.pdf}

\usepackage[cmex10]{amsmath}
\usepackage{amssymb}
\usepackage{mathtools}
\usepackage{amsfonts}

\usepackage{subfig}

\usepackage{algorithmicx}
\usepackage{algpseudocode}
\usepackage[ruled]{algorithm}
\definecolor{light-gray}{gray}{0.75}
\algrenewcommand{\algorithmiccomment}[1]{\hskip3em{{\footnotesize \textcolor{light-gray}{$\blacktriangleright$}}} #1}
\usepackage{multirow} 
\usepackage{rotating} 
\usepackage{booktabs} 
\usepackage{colortbl} 
\usepackage{tablefootnote} 

\usepackage[pdftex,colorlinks=true,urlcolor=blue,citecolor=blue]{hyperref}

\usepackage{xspace}

\let\labelindent\relax
\usepackage{enumitem}

\usepackage{blindtext}

\begin{document}
\date{}


\title{
Towards Perception-based Collision Avoidance for UAVs when Guiding the Visually Impaired
}

\author{Suman Raj, Swapnil Padhi, Ruchi Bhoot, \\Prince Modi and Yogesh Simmhan\\
Department of Computational and Data Sciences, \\Indian Institute of Science, Bangalore 560012 India\\
Email: \{sumanraj, simmhan\}@iisc.ac.in
}

\maketitle

\begin{abstract}
Autonomous navigation by drones using on-board sensors combined with machine learning and computer vision algorithms is impacting a number of domains, including agriculture, logistics and disaster management. In this paper, we examine the use of drones for assisting visually impaired people (VIPs) in navigating through outdoor urban environments. Specifically, we present a perception-based path planning system for local planning around the neighborhood of the VIP, integrated with a global planner based on GPS and maps for coarse planning. We represent the problem using a geometric formulation and propose a multi-DNN based framework for obstacle avoidance of the UAV as well as the VIP. Our evaluations conducted on a drone--human system in a university campus environment verifies the feasibility of our algorithms in three scenarios: when the VIP walks on a footpath, near parked vehicles, and in a crowded street. 
\end{abstract}


\section{Introduction}
There is a growing trend for fleets of Unmanned Aerial Vehicles (UAVs), also called drones, being used for food delivery, flying ambulances~\cite{drone-first-aid}, urban safety monitoring, and disaster response~\cite{ollero2004motion}. UAVs have become popular due to their agility in urban spaces and their lightweight structure. Further, recent advancements in Deep Learning (DL) and Computer Vision (CV) algorithms have helped to hasten their autonomous navigation and operations. In this paper, we focus on a novel use case of developing a scalable platform for UAVs that help \textit{Visually Impaired People (VIPs)} to lead an active urban lifestyle~\cite{suman2023chi}. 

Autonomous navigation of drones is made possible by the integration of sensors such as GPS, cameras, and accelerometers that help the drone to perceive and understand its environment. In particular, video streams from the camera on-board the drone when coupled with Deep Neural Networks (DNN) models, CV algorithms and advanced control logic allow drones to achieve situation awareness and take in-flight decisions, such as avoiding obstacles or changing course to reach a specific destination. Lastly, advances in GPU edge accelerators and 4G/5G communications make it feasible to perform such real-time inferencing over video streams using onboard edge accelerators or remote Cloud servers~\cite{sumanccgrid}. 

\begin{figure}
  \centering
	\includegraphics[width=0.9\columnwidth]{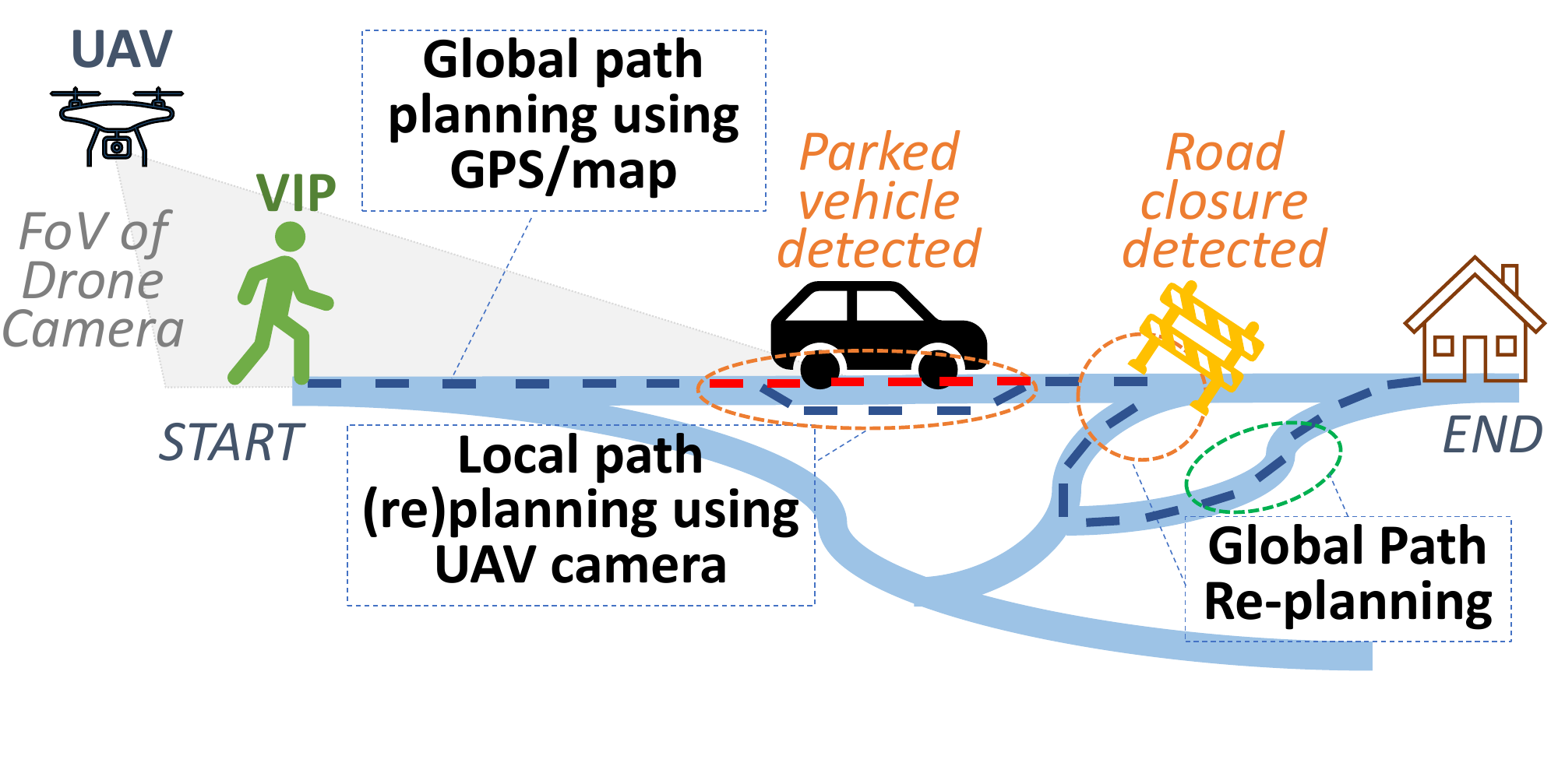}
    \caption{Global and Local Path Planning for VIPs using UAV}
    \label{fig:path-planning}
    \vspace{-0.1in}
\end{figure}

In this paper, we focus on how a UAV can serve as a ``buddy drone'' to autonomously guide a VIP through an urban environment to a destination location, while avoiding obstacles along the streets. This requires both navigation assistance and obstacle-avoidance along the planned path, and the primary sensing modality is limited to just the video stream from the drone.
As illustrated in Fig.~\ref{fig:path-planning}, we propose a dual-layer path planning framework: (1) A coarse-grained global plan is generated based on the shortest distance along a city map known \textit{a priori}, and (2) A fine-grained local plan is generated based on the video feed around the VIP collected by the UAV, and the local path updated when obstructions are detected along the route; road closures may trigger a re-planning of the global path.

\textbf{Challenges.} Since this is a time-sensitive application, there are multiple challenges we need to address: 
\begin{itemize}
    \item We need a framework to generate a coarse-grained path with minimal inputs, and generate a fine-grained path as the VIP progresses on their path. 
    \item We need a methodology for obstacle avoidance -- for the VIP as well as for the drone -- while ensuring that the VIP always stays in the Field of View (FoV) of the UAV.
    \item Given the complexity of the decision making, we need a stable and robust compute platform that can run multiple and diverse DNN models for real-time inferencing of the UAV's visual feeds.
\end{itemize}

\textbf{Contributions.}
The key contributions of this paper are:
\begin{enumerate}
    \item We motivate a cyber-physical-social scenario of guiding VIPs using buddy drones from a start point to a destination using live visual feeds and audio cues.
    \item We propose a two-layered path planning strategy for obstacle avoidance of the VIP and the UAV -- namely the \textit{Global planning} using Global Positioning System (GPS) coordinates and map information, and the \textit{Local planning} using live visual feed from the UAV (\S~\ref{sec:framework}). 
    \item We design a detailed framework for local path planning and obstacle avoidance. This is modeled using a geometric formulation, combined with inferencing over multiple DNNs for object detection and depth estimation to determine the geometric variables and decisions (\S~\ref{sec:geometric-formulation},~\ref{sec:local-plan}).
    \item We evaluate our proposed framework using specific UAV--human interaction scenarios in an outdoor environment -- when the VIP walks on a footpath, near parked vehicles, and in a crowded street -- to illustrate the effectiveness of our techniques (\S~\ref{sec:experiments}).
\end{enumerate}

Besides these, we also discuss the related work in \S~\ref{sec:related-work} and offer our conclusions in \S~\ref{sec:conclusion}.

\section{Related Work}\label{sec:related-work}
\subsection{Drones-Based Assistive Solutions for VIPs}
The use of drones to assist VIPs with their navigation has gained popularity. Nasralla, et al.~\cite{nasralla2019computer} study the potential for employing deep learning and computer vision-enabled UAVs to aid VIPs in a smart city. The use of drones for VIP's navigation is explored by Avila, et al.~\cite{avila2015dronenavigator,avila2017dronenavigator}. In their study, Zayer, et al.~\cite{al2016exploring} investigate how well blind runners could use drones to guide them using the sound of rotors.

However, these works have been accomplished in a highly controlled setting, and have not been designed or validated for a more complex metropolitan outdoor environment, which we aim to do. Our prior work, Ocularone~\cite{suman2023chi}, proposes the hig-level ideas for a platform to coordinate and manage a heterogeneous UAV feet that guide VIPs. A key design element was detecting the safety of the VIP using computer vision models to detect falls and take visual cues using hand gestures. This paper extends those ideas to actively guide the navigation of the VIP, in particular through perception and avoidance of local obstacles.

\subsection{Vision-based Real-Time Obstacle Avoidance}
Algorithms over images and videos have been used for collision avoidance.
Early works such as Watanabe, et al.~\cite{watanabe2007vision} use a single 2D passive vision sensor and extended Kalman filter (EKF) to estimate the relative position of obstacles using a collision-cone approach. Recent advances in machine perception, deep learning and GPU-accelerated computing have led to a large body of literature that integrate computer vision-based approaches with deep learning~\cite{5651028}. We leverage some of these advances in our work.

Kaff, et al.~\cite{al2017obstacle} propose an obstacle detection algorithm for monocular camera in UAVs by analyzing the change in the pixel-area of the approaching obstacles. The avoidance is maneuvered by estimating the obstacle's 2D position in the image and combining it with tracked waypoints. Lee, et al.~\cite{lee2021deep} use deep learning models for monocular obstacle avoidance for UAVs that are flying in tree plantation. They classify obstacles as critical or low priority based on the height of the bounding boxes of the detected obstacles (trees). We use similar technique of calculating the tentative heading of the VIP using the maximum width of free space. We additionally integrate our logic with pixel density in the frames. 

\subsection{Real-Time Dynamic Target Tracking}
A key requirement for assisting the VIP is to for the drone to detect and track them using its cameras.
Woods, et al.~\cite{woods2015dynamic} highlight the benefit of using potential field navigation to follow a dynamic human target by a drone and for obstacle avoidance. However, they require an external motion capture system, which limits its deployment in outdoor environments. Xia, et al.~\cite{xia2011human} propose a tracking algorithm based on a segmentation scheme that uses depth images to segment humans and extract the contours of the figure. But it is highly dependent on the accurate detection of the person's head, which may be occluded when walking in public spaces. Instead, we have the VIP wear a hazard vest and detect a major portion of the VIP's body, which is more robust. 

Karlsson, et al.~\cite{karlsson2021monocular} use a Kalman filter combined with object detection and classification for real-time tracking of objects for collision avoidance. They translate the pixel coordinates of the object into heading commands using trigonometric operations. This motivates the geometric formulation used by us. They, however, limit their validation to indoor environments with only one (pedestrian) obstacle. We consider multiple stationary obstacles in an outdoor environment.

\subsection{Static and Dynamic Path Planning}
In order to guide the VIP through a pre-planned path, we leverage the concept of a two-phase path planner namely, local and global as proposed by Far Planner~\cite{yang2022far}. Far Planner updates a global visibility graph where the method eliminates visibility edges blocked by the dynamic obstacles and later on reconnects the edges after regaining the visibility. We similarly use a global graph which updates based on the visual feeds. Kusnur, et al.~\cite{kusnur2021search} solve a path planning problem for robot and sensor trajectories that maximize information gain in such tasks. This is helpful for our problem formulation where the camera orientation and FoV impacts the coverage of VIP and surroundings in the frame. 

We have designed a prototype in our earlier work~\cite{suman2023chi}, where the drone autonomously follows a VIP using a $PD$ control loop algorithm, along with real-time situation awareness and gesture recognition. In this work, we focus majorly on the vision-based navigation not only for the UAV, but also for tracking and guiding the VIP through obstacles in an outdoor environment. 

\section{Formulation for Local Planning}\label{sec:geometric-formulation}
We address the problem of path planning and navigation assistance for VIPs from a source to a destination location as a two-phase problem -- global planning of the route at the gross level based on classical map and GPS techniques, and then local planning for the drone to assist the VIP in walking along the global path and responding to fine-grained local conditions such as obstacles. Here, we discuss a geometric model for assigning distance constraints between the drone and the VIP when performing local planning.

\begin{figure}
  \centering
	\includegraphics[width=0.7\columnwidth]{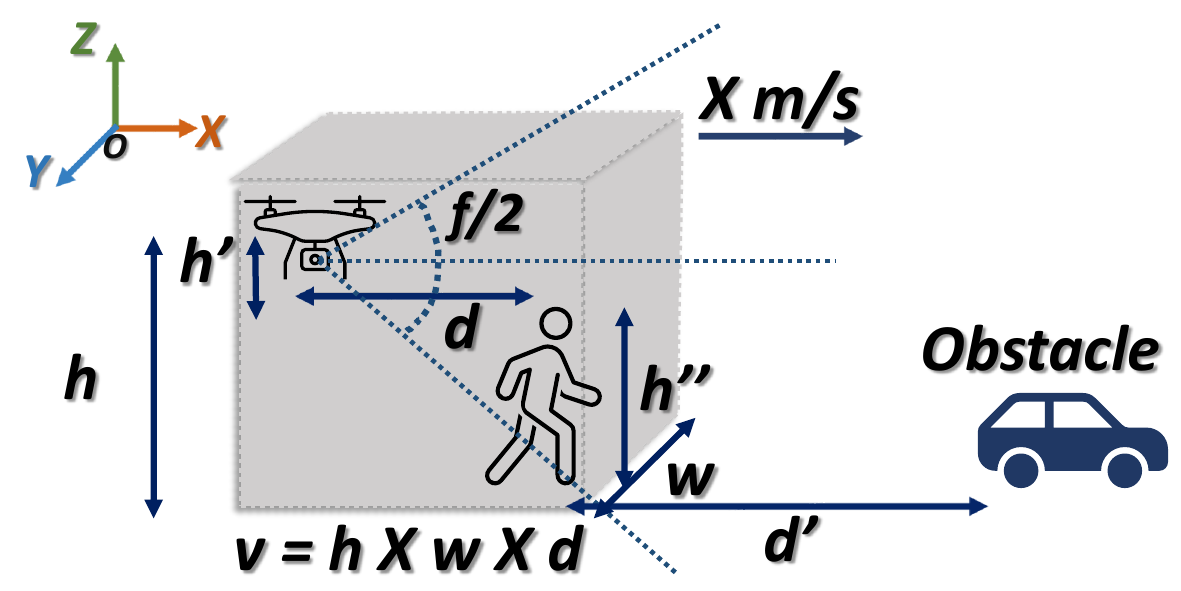}
    \caption{Problem Formulation of Local Path Planning}
    \label{fig:local-path-planning}
\end{figure}

We consider a 3D volume with dimensions $h,d,w$, where $h$ is the height of the drone from the ground, $d$ is the logitudinal distance of the drone from the VIP, and $w$ is the width of lateral free space that the VIP requires for unobstructed motion (Fig.~\ref{fig:local-path-planning}). $h'$ is the vertical offset between the top of the VIP with height $h''$ and the current height of the drone. For simplicity, we assume that the orientation of the drone with respect to the VIP is fixed, and the heading of the VIP coincides with the heading of the drone. While a quadrotor UAV has \textit{six} degrees-of-freedom (DoF), we consider only \textit{four} -- translation in X, Y and Z axes, and orientation about the Z axis (yaw).

\begin{figure*}
  \centering
	\includegraphics[width=0.8\textwidth]{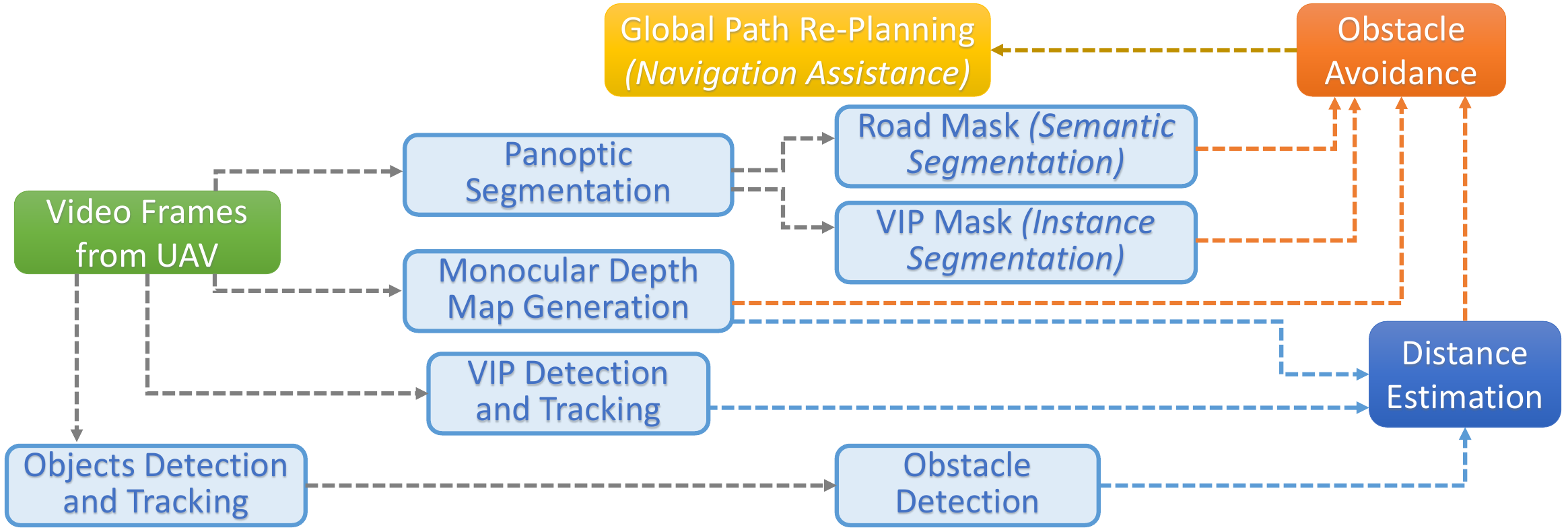}
    \caption{Path Planning Framework }
    \label{fig:methodology}
\end{figure*}

\subsection{Relative Positioning of the Drone}

The height offset ($h'$) and the distance ($d$) of the drone relative to the VIP is chosen such that the VIP is always in the FoV of the drone's onboard camera. If $f$ is the FoV of the camera, we can ensure that at least the top of the VIP's head is visible to the drone by setting the offset and height as:
\begin{equation}\label{eqn:tan}
    \tan{(\frac{f}{2})} = \frac{h'}{d}
\end{equation}
This however is not adequate for accurate tracking of the VIP. Instead, we define a range of valid values for $h' \in [h'',h_{max}]$ and $d \in [d_{min},d_{max}]$, such that we have a straight line with coordinates $(h'',d_{max}),(h_{max},d_{min})$ satisfying equation~\ref{eqn:tan}. We set $h_{max}$ and $d_{min}$ as the point on the straight line where roughly \textit{two-thirds} of the person starting from head is visible to the drone. We consider $d_{max}$ as the farthest distance the drone can go behind the VIP such that it can detect obstacles upto a minimum distance $d'$ ahead of the VIP, as defined by equation~\ref{eqn:d}. 

For practical purposes, we bound the values of $d_{min} \geq 1~m$ and $d_{max} \leq 10~m$ to stay within safe proximity limits.

\subsection{Visibility Ahead of the VIP to Detect Obstacles}
$d'$ is the \textit{safety distance} on the path directly ahead of the VIP that needs to be obstacle-free while they navigate. While this depends on multiple factors, we consider two significant ones based on the need for the VIP to react when an obstacle is detected. The first is the \textit{detection time} ($t_{detect}$ in seconds) of the obstacle, between the camera observing it in its video frame to the processing of the frame using DNN models to identify the obstacle. The second is the \textit{human reaction time} ($t_{react}$ in seconds), which is the average time required for the VIP to comprehend and respond to an audio/sensory cue from our system when notifying them of an obstacle. This depends on the individual, but has been observed to be longer for the visually impaired~\cite{wood1994effect}. Let the speed at which the VIP is walking be $x~m/s$. Then, we estimate $d'$ as:
\begin{equation}\label{eqn:d}
    d' = x \cdot (t_{detect} + t_{react})
\end{equation}
i.e., we should have a safety distance such that we can digitally detect an obstacle, inform the VIP and they have time to respond to the situation, at the current speed of walk.
Besides this, we can add a safety margin distance ($d_{buffer}$), which we set to be $0.05 \cdot d'$ in our experiments. Hence, the final position of the drone behind the VIP is set such that it can have a view of $(d + d' + d_{buffer})$ ahead of it. 


\section{Local Path Planning using DNN Models}\label{sec:local-plan}

Next, we propose an approach for local planning along the global route, which involves tracking of the VIP by the drone, detection of obstacles and avoidance of obstacles. The avoidance may through local cues to the VIP, such as going around a parked vehicle, or trigger a global re-planning of the route, say, due to road closures. Local planning uses a collection of contemporary DNN models, which operate on the UAV's video feeds to bootstrap the obstacle avoidance algorithm. The high-level decision flow is shown in Fig.~\ref{fig:methodology}, and described in detail below.

\subsection{Distance Estimation} 
Classifying any object in front of the VIP as an obstacle requires us to estimate the distance between the VIP and each object in the frame. We concurrently run three DNN models over the video streams to help with \textit{distance estimation}.  
\begin{itemize}
    \item \textit{VIP Detection}: The UAV has to track the VIP and follow them at a distance. This requires the drone to first identify them within the public space, and orient itself. We assume that the VIP wears a distinctive accessory, like a vest, cap or Hiro tag~\cite{colley2020exploring}, which serves as a unique and robust visual marker. Once detected, we also need to determine the orientation of the VIP with respect to the drone, i.e., is the drone is facing the front or the rear of the VIP. We retrain the popular off-the-shelf YOLOv8 DNN model to specifically detect a VIP wearing a hazard vest with real-time inferencing.
    \item \textit{Obstacle Detection}: We need to identify objects ahead of the VIP and classify them as obstacles. We again use YOLOv8 to detect objects in the frame, which returns the classes of the objects and their bounding box coordinates in the frame. The classes are further mapped as obstacles in the outdoor environment, such as car, bicycle, humans, etc.
    \item \textit{VIP and Obstacles Tracking}: Once the framework detects the VIP and the obstacles, we use the ByteTrack algorithm~\cite{zhang2022bytetrack} for multi-object tracking (MOT), which returns a \textit{unique identifier} and \textit{class type} for each object in the frame. This improves robustness of objects detection by leveraging recent historic data in the video feed and helps predict the bounding box coordinates even if there are missed detections. This helps us estimate the rate at which the VIP is approaching an obstacles and helps in timely collision avoidance.
    \item \textit{Monocular Depth Map Generation}: Estimating the distance to obstacles using geometrical models is challenging for complex navigation scenarios in public spaces~\cite{lee2021deep}. Instead, we use the MiDas~\cite{ranftl2020towards} deep learning model to generate a monocular depth map over the video frames. The frame is shaded with hues for pixels based on the distance of the object representing a pixel from the camera. This uses \textit{zero-shot cross-dataset} transfer performance, which is a more faithful proxy for ``real-world'' performance. We reduce the computational complexity of the problem by converting the $3$ channel RGB image to grayscale monochrome and then use the value of the hues.
\end{itemize}

Depth maps provide us with relative depth estimates of detected objects (REV) in the frame. We frame a supervised model using a second-order polynomial regression. We have collected datasets using a laster pointer with $1~cm$ precision using reference objects belonging to COCO classes, i.e. person, car, bicycle, etc. We train our model and achieve a Root Mean Square Error (RMSE) of $1.2~m$. 

\begin{algorithm}[!t]
\caption{Obstacle Avoidance Algorithm}\label{algo:obstacle-avoidance-enhanced}
\begin{algorithmic}[1]

\Function{Obstacle Avoidance}{VIP and object mask, VIP and object detection}
\State Exclude pixel for masks
\State Generate depth map
\For{Each partition in frame}
\State Calculate H(i) using (Eqn.~\ref{eqn:pixelsum})
\EndFor
\State Sort H
\For {each partition in H}
\If{partition contains objects}
\State Max width(free space (partition)) $<$ threshold */no space in partition for VIP to move ahead*/
\State reject this partition and continue
\Else 
\State Set the heading towards max width 
\EndIf
\EndFor
\State Invoke rerouting of global path
\EndFunction
\end{algorithmic}
\end{algorithm}

\begin{algorithm}[!t]
\caption{Free Space Detection}\label{algo:free-space}
\begin{algorithmic}[1]
\Function{Free Space}{VIP and object detection,frame}
\State Free space coordinates, $C = []$
\For{Detected objects within d'}
\State Sort detections across x-axis in frame
\State Append x-coordinates between two objects
\EndFor
\State $P[]$ = $n$ vertical parts(frame)
\For{$i$ in $C$}
\If{$C[i]$ and $C[i+1]$ in same partition, $P_j$}
\State Free space width, $W = C[i+1][x_1] - C[i][x_2]$
\ElsIf{$C[i]$ and $C[i+1]$ overlap}
\State Retain $C[i]$ as starting box
\State Increment $i$ until overlapping ends
\State $W_{j} = C[i+n][x_1] - C[i][x_2]$
\ElsIf{$C[i] \in P_j$, $C[i+1] \in P_{j+1}$} 

*/free space spans partitions*/
\State $W_j$ = end($P_j$) - $C[i][x_2]$ 
\State $W_{j+1}$ = $C[i+1][x_1]$ - start($P_{j+1}$)
\Else 

*/free space spans multiple partitions*/

\State $W_j$ = end($P_j$) - start($P_j$)
\EndIf
\EndFor
\State Return C, W for each partition 
\EndFunction
\end{algorithmic}
\end{algorithm}

\begin{algorithm}[!t]
\caption{Road Edge Detection Algorithm}\label{algo:road-edge-detection}
\begin{algorithmic}[1]

\Function{Detect Road Edges}{VIP detection}
\State Set $90\times90$ pixel box on lower left and right of VIP detection box

*/ $90\times90$ corresponds to $\approx0.5~m$*/
\For{each pixel in pixel box}
\State Slice object segmentation mask
\If{Road detected}
\State pixel value = 255
\Else
\State pixel value = 0
\EndIf
\State Mean, M = mean(pixel values)
\If{M $>$ threshold}
\State VIP is at a safe distance from edge
\Else
\State Issue warnings to move away from edge
\EndIf
\EndFor
\EndFunction
\end{algorithmic}
\end{algorithm}

\subsection{Obstacle Avoidance}

We design a robust obstacle avoidance module that operates on processed image frames with estimated absolute distances. The module segments the input frame into $n$ equally-spaced vertical partitions, where $n$ is an odd number to enable the VIP (Visually Impaired Person) to maintain a central heading at $0^\circ$. The extreme partitions help detect obstacles such as walls or road edges, which are critical in narrow path scenarios. In our formulation, we set $n=3$ for computational efficiency while ensuring sufficient spatial resolution for navigation.

For each partition $i$, we compute the mean depth value, $H(i)$, as shown in Equation~\ref{eqn:pixelsum}:

\begin{equation}\label{eqn:pixelsum}
H(i) = \frac{\sum(value(pixels))}{\sum(pixels)}
\end{equation}

This value represents the average distance to obstacles within the partition. The obstacle avoidance logic is detailed in Algorithm~\ref{algo:obstacle-avoidance-enhanced}. First, pixel contributions from the VIP mask are excluded (line 2), as the VIP should not be treated as an obstacle. The depth map is then generated for the current frame. We sort partitions based on $H(i)$ in increasing order, favoring those with fewer nearby obstacles.

Next, we assess whether sufficient navigable space exists for the VIP by applying the \textit{Free Space Detection} algorithm (Algorithm~\ref{algo:free-space}). This algorithm analyzes gaps between detected objects (excluding the VIP) along the x-axis of the frame, calculates the free-space widths ($W_j$) within and across partitions, and identifies partitions that offer safe paths for movement. Partitions with width below a defined threshold are rejected (Algorithm~\ref{algo:obstacle-avoidance-enhanced}, line 9), as they cannot accommodate the VIP.

In the event that no suitable local path is found, a re-routing of the global plan is triggered (\S~\ref{sec:framework}). Otherwise, the VIP is directed towards the partition offering maximum safe free-space width and minimal average depth, effectively guiding them away from obstacles.

An additional safety layer is implemented through the \textit{Road Edge Detection} algorithm (Algorithm~\ref{algo:road-edge-detection}). A $90\times90$ pixel box is overlaid on the lower-left and lower-right regions of the VIP detection box (corresponding to approximately $0.5~m$ in the real world). By analyzing pixel values in these regions, the algorithm determines the VIP’s proximity to road edges. If the mean pixel value within this region falls below a threshold—indicating potential proximity to an edge—an auditory or haptic warning is issued to prompt corrective action.

To extract object and road regions for all these processes, we employ panoptic segmentation using Detectron2~\cite{wu2019detectron2}, combining semantic segmentation to identify road and pavement areas with instance segmentation to uniquely identify the VIP and other objects.

Overall, this obstacle avoidance module ensures safe navigation by continuously evaluating available free space and road context in real-time, allowing the VIP to navigate dynamic outdoor environments with enhanced confidence and safety.

\begin{figure}[t]
\centering
  \subfloat[Initial generated path]{
    \includegraphics[width=0.45\columnwidth]{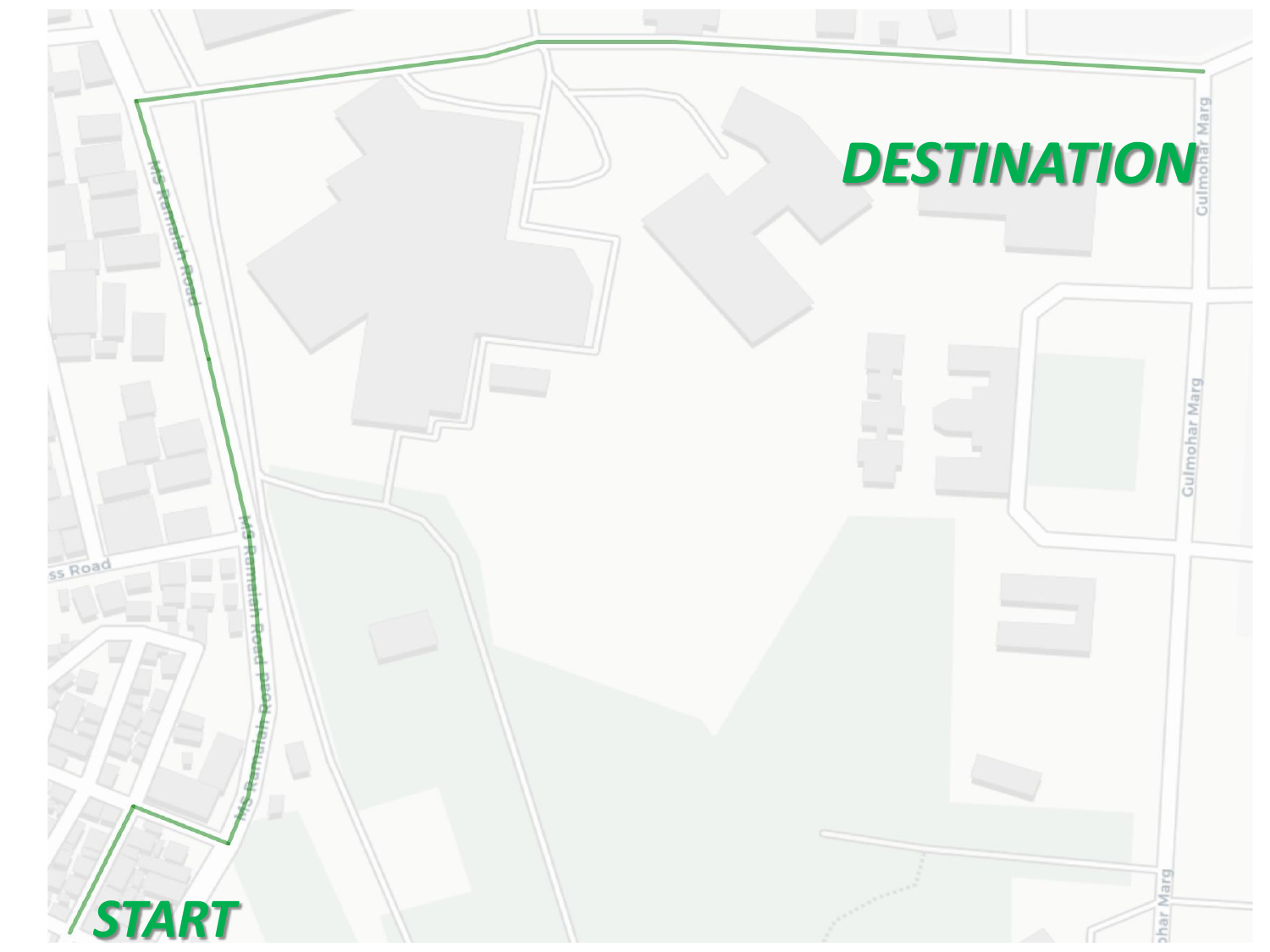}
   \label{fig:initial-path}
  }
  \subfloat[Path Re-routing]{
   \includegraphics[width=0.45\columnwidth]{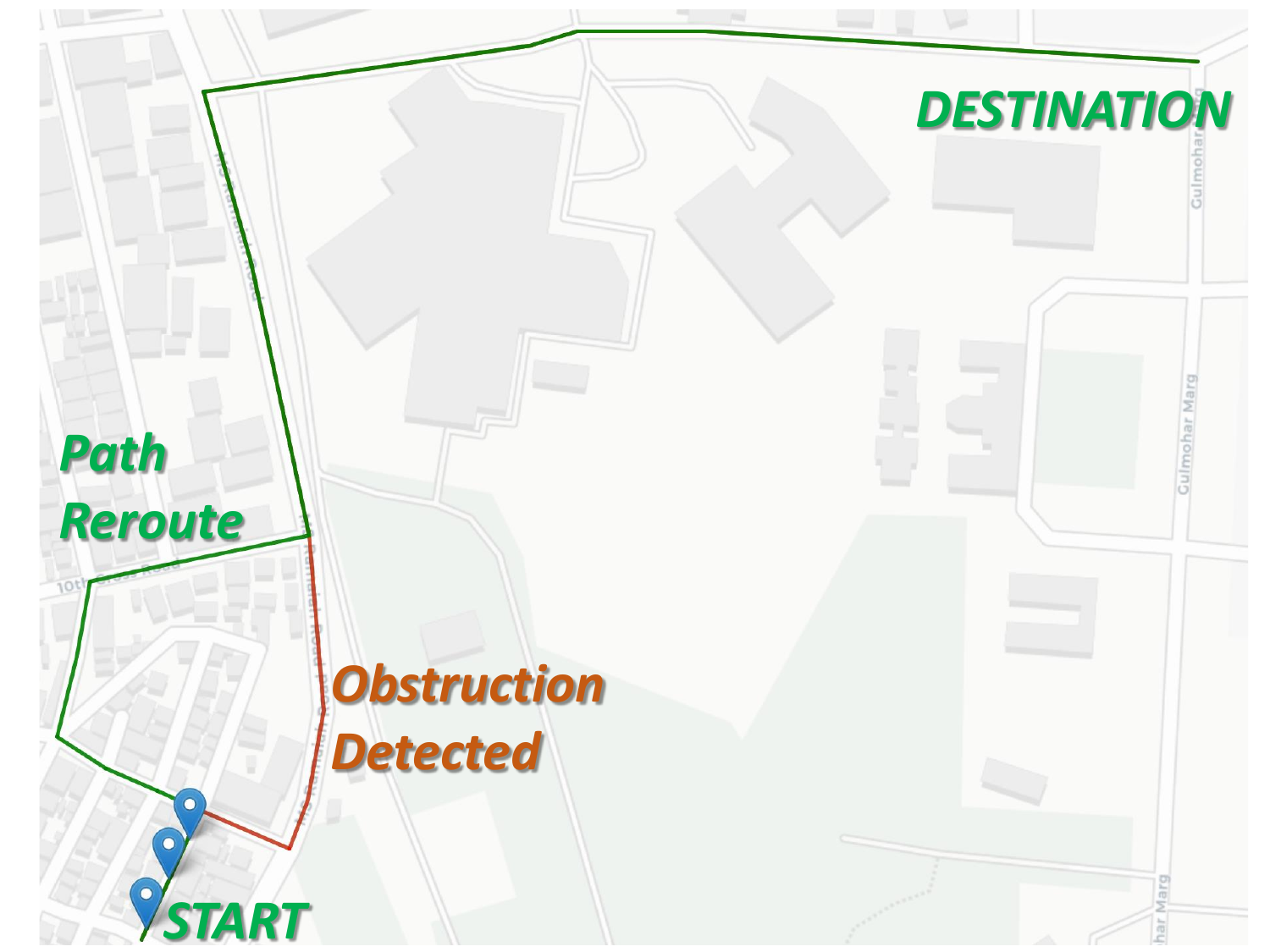}
    \label{fig:final-path}
  }
\caption{Static and Dynamic Path Planning}
\label{fig:initial-final-path}
\end{figure}

\begin{figure*}[t]
\centering
  \subfloat[Input image]{
    \includegraphics[width=0.3\columnwidth]{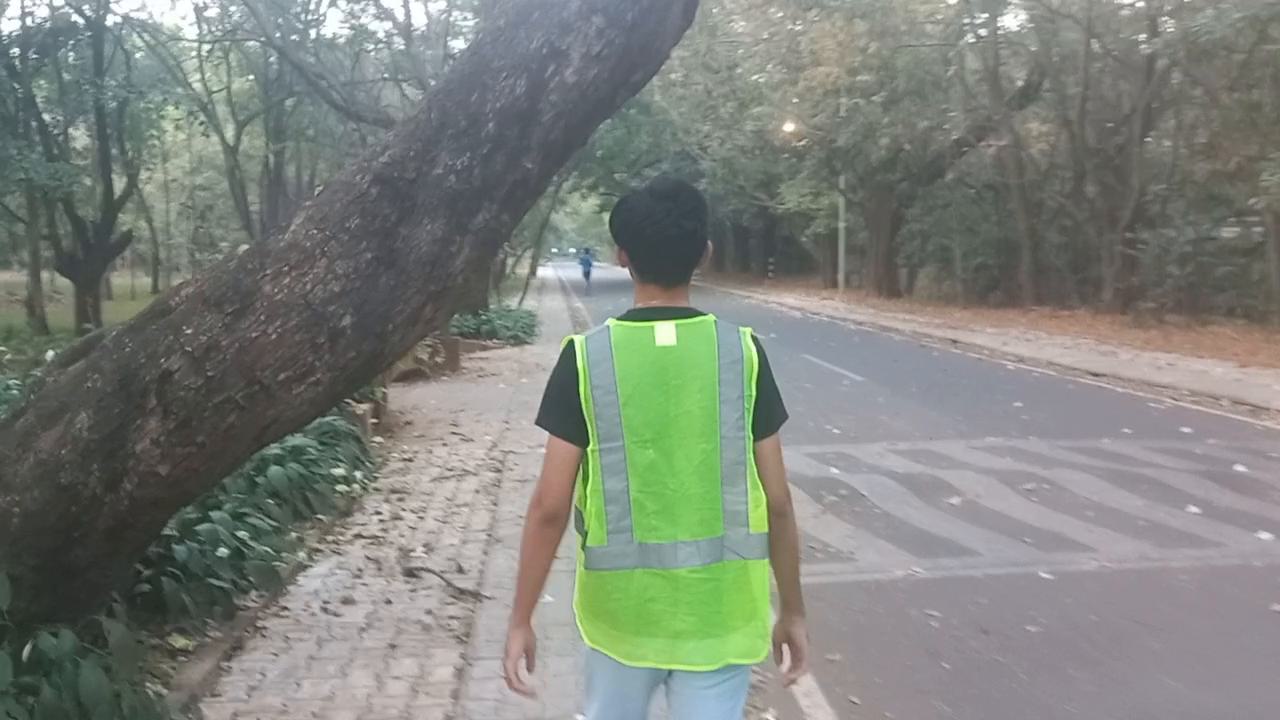}
   \label{fig:1a}
  }
  \subfloat[Object detection]{
   \includegraphics[width=0.3\columnwidth]{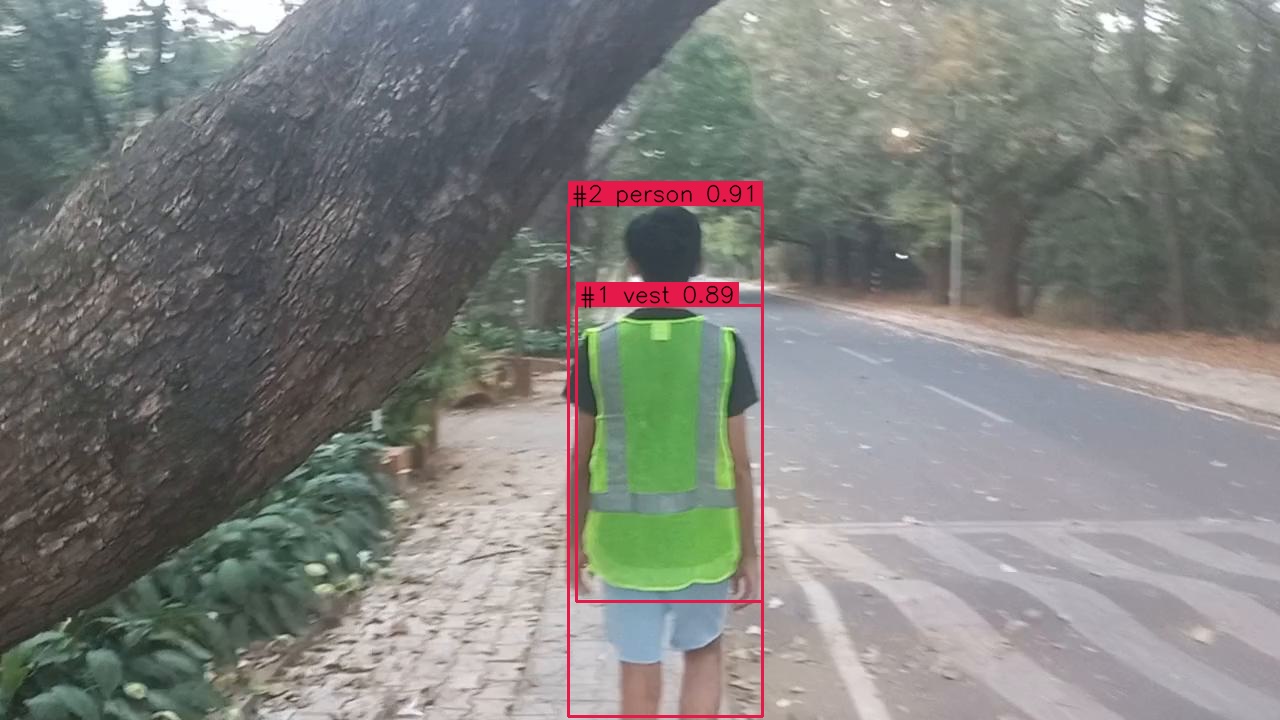}
    \label{fig:1b}
  }
    \subfloat[Road segmentation]{
   \includegraphics[width=0.3\columnwidth,height=2.15cm]{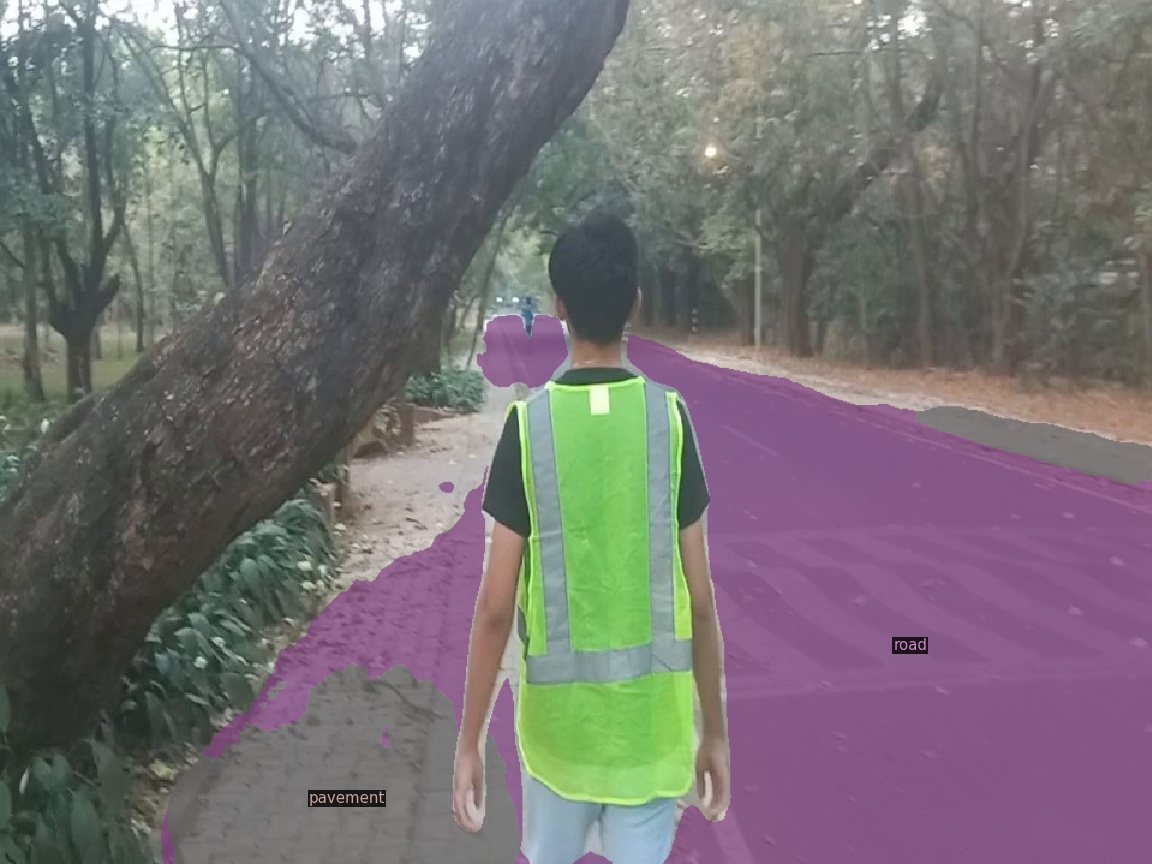}
    \label{fig:1c}
  }\\
   \subfloat[VIP segmentation]{
   \includegraphics[width=0.3\columnwidth]{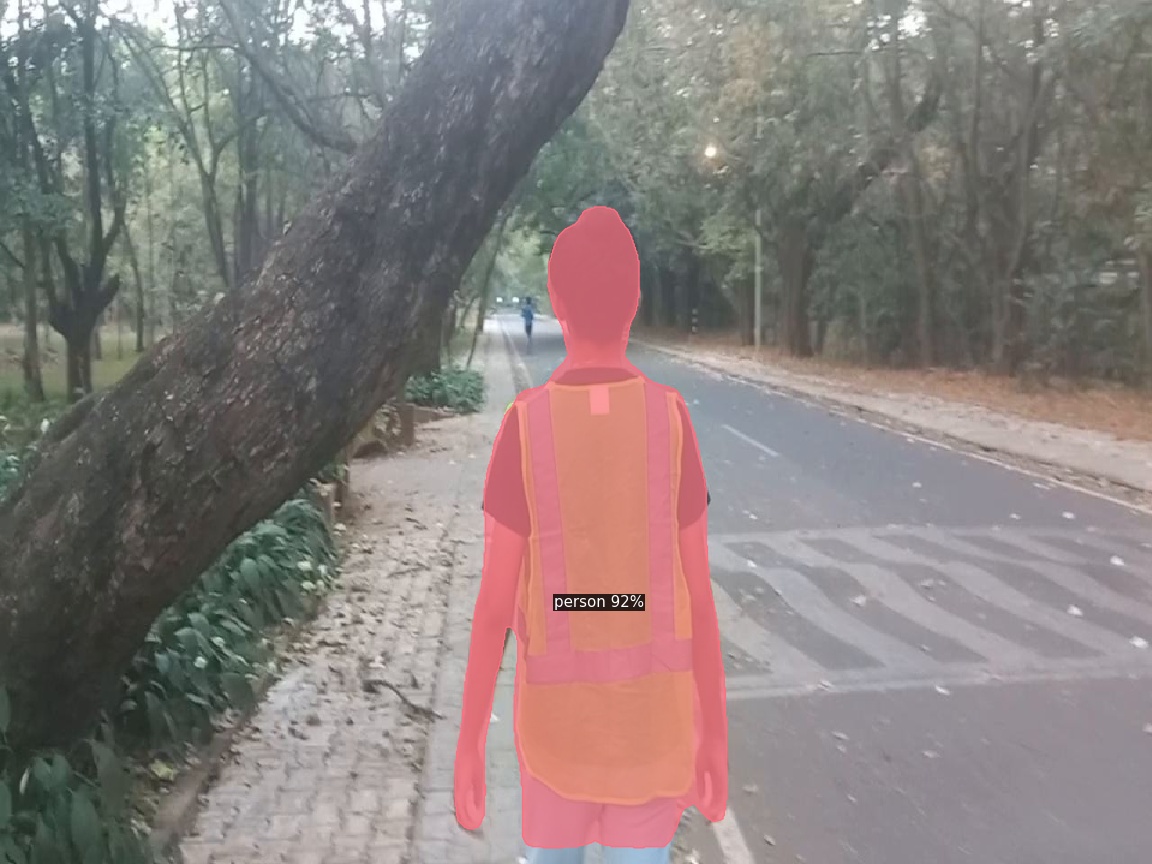}
    \label{fig:1d}
  }
  \subfloat[Depth map]{
   \includegraphics[width=0.3\columnwidth]{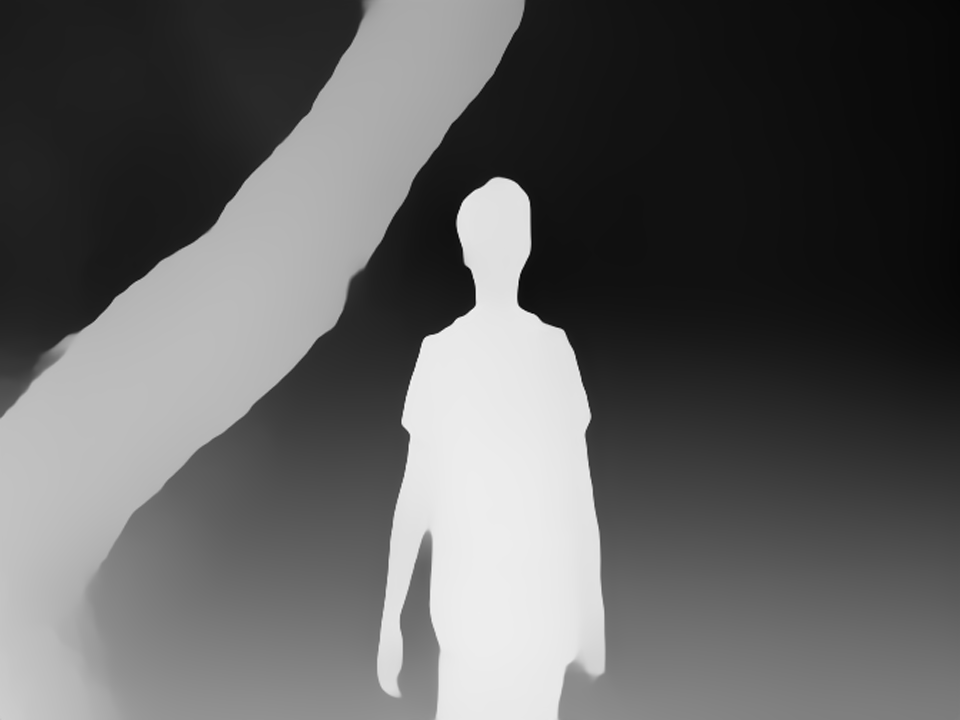}
    \label{fig:1e}
  }
  \subfloat[Required heading]{
   \includegraphics[width=0.3\columnwidth]{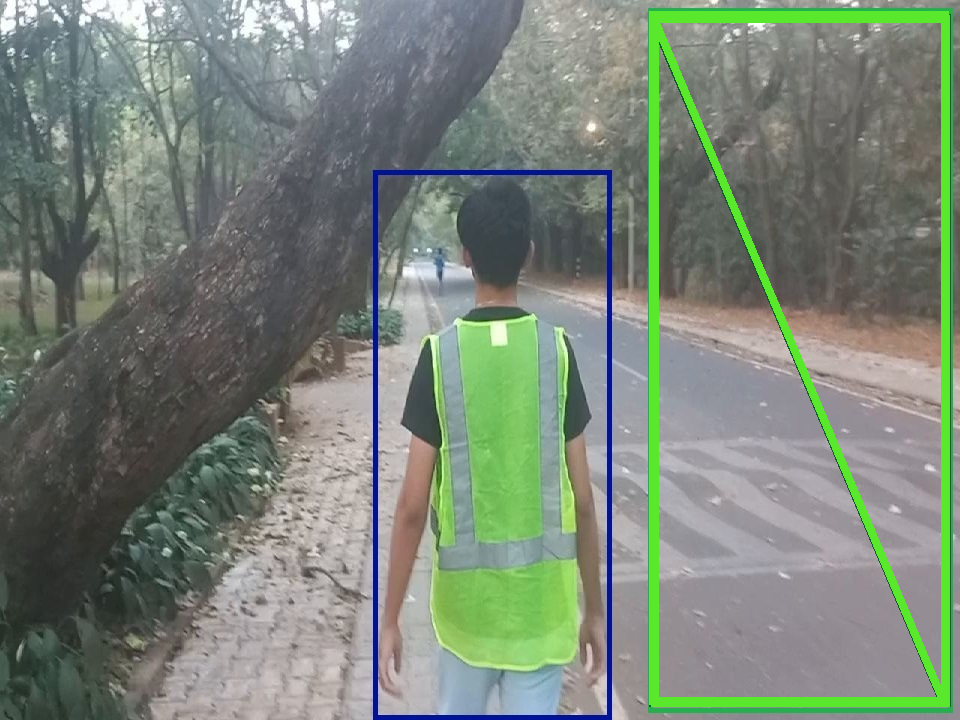}
    \label{fig:1f}
  }  
\caption{Obstacle avoidance along a pedestrian footpath}
\label{fig:scenario-1}
\end{figure*}

\begin{figure*}[t]
\centering
  \subfloat[Input image]{
    \includegraphics[width=0.3\columnwidth]{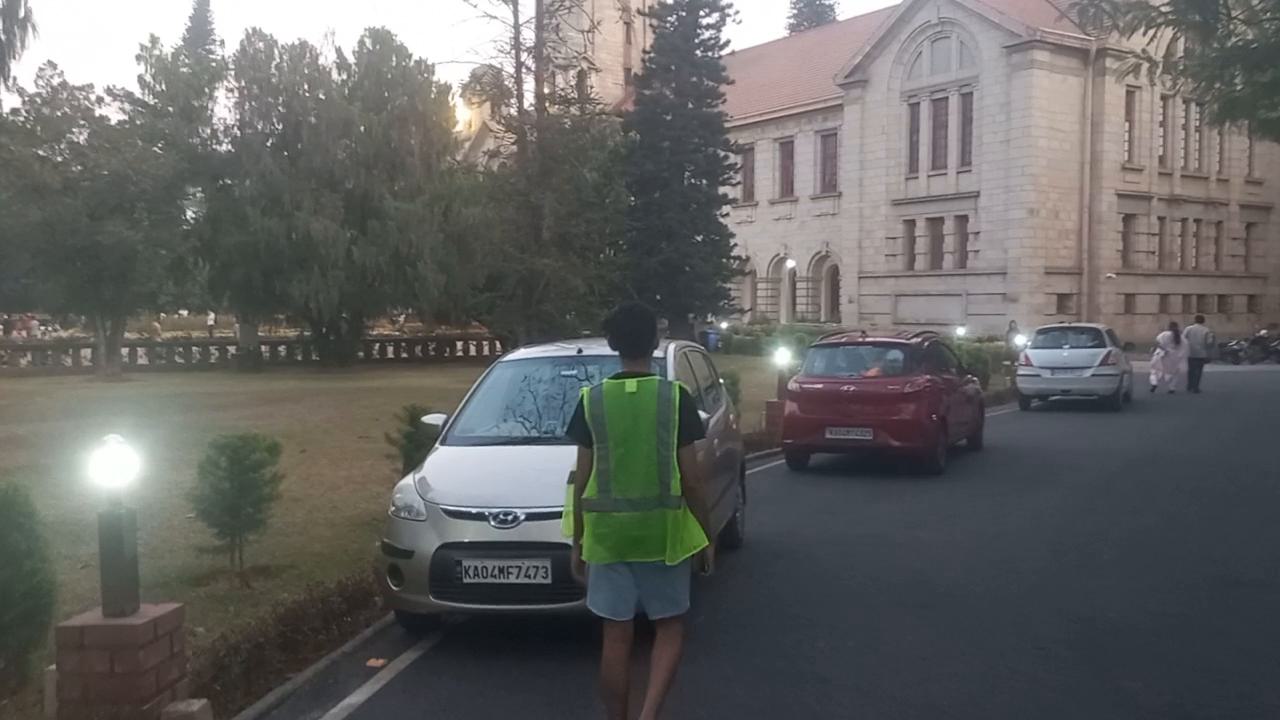}
   \label{fig:2a}
  }
  \subfloat[Object detection]{
   \includegraphics[width=0.3\columnwidth]{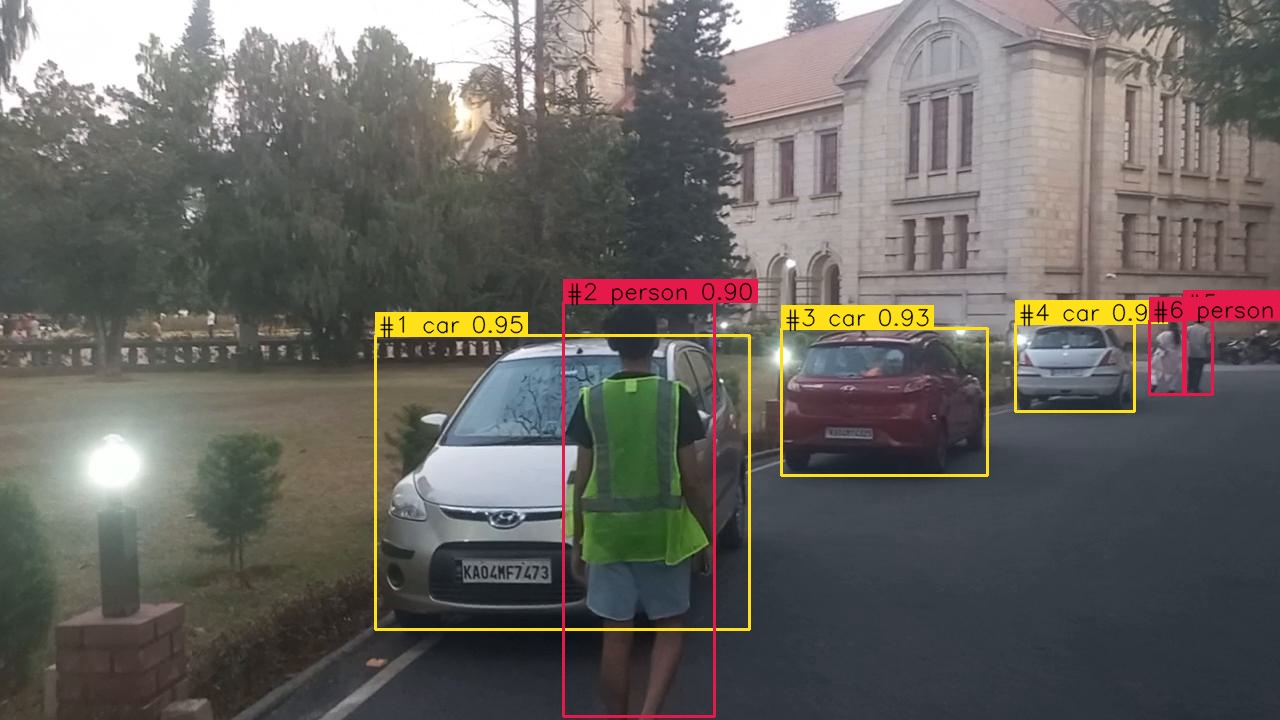}
    \label{fig:2b}
  }
    \subfloat[Road segmentation]{
   \includegraphics[width=0.3\columnwidth,height=2.15cm]{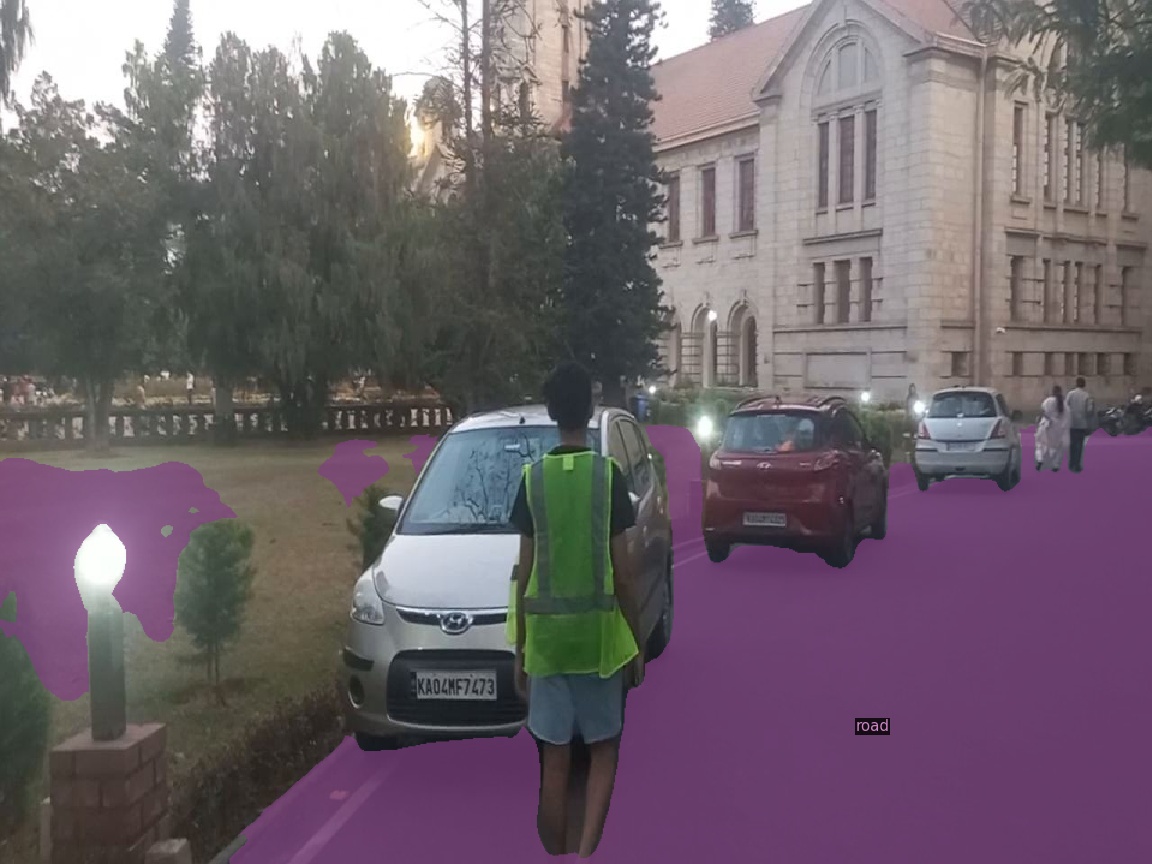}
    \label{fig:2c}
  }\\
   \subfloat[VIP segmentation]{
   \includegraphics[width=0.3\columnwidth]{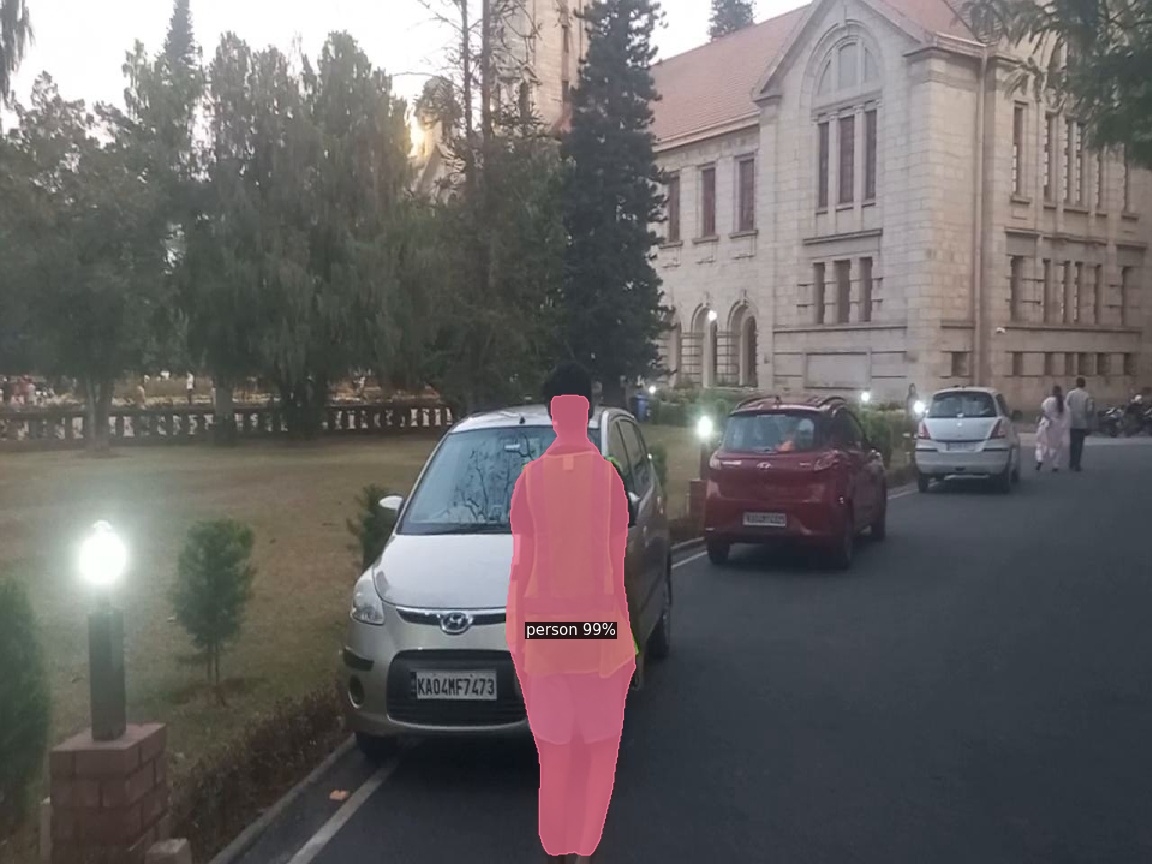}
    \label{fig:2d}
  }
  \subfloat[Depth map]{
   \includegraphics[width=0.3\columnwidth]{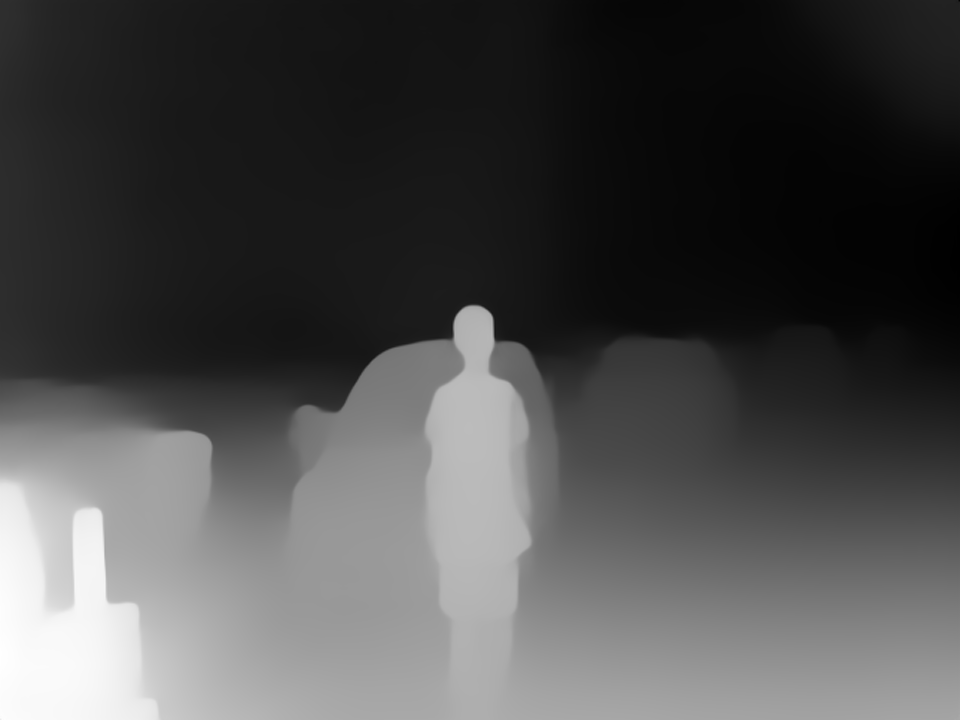}
    \label{fig:2e}
  }
  \subfloat[Required heading]{
   \includegraphics[width=0.3\columnwidth]{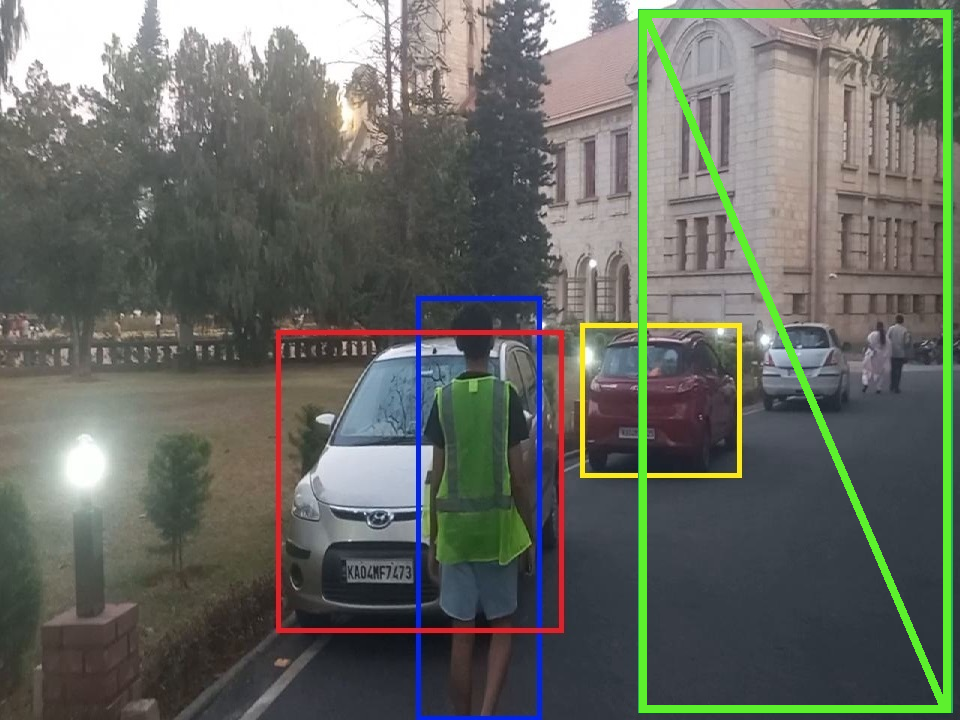}
    \label{fig:2f}
  }  
\caption{Obstacle avoidance in case of parked vehicles along the street}
\label{fig:scenario-2}
\end{figure*}

\begin{figure*}[t]
\centering
  \subfloat[Input image]{
    \includegraphics[width=0.3\columnwidth]{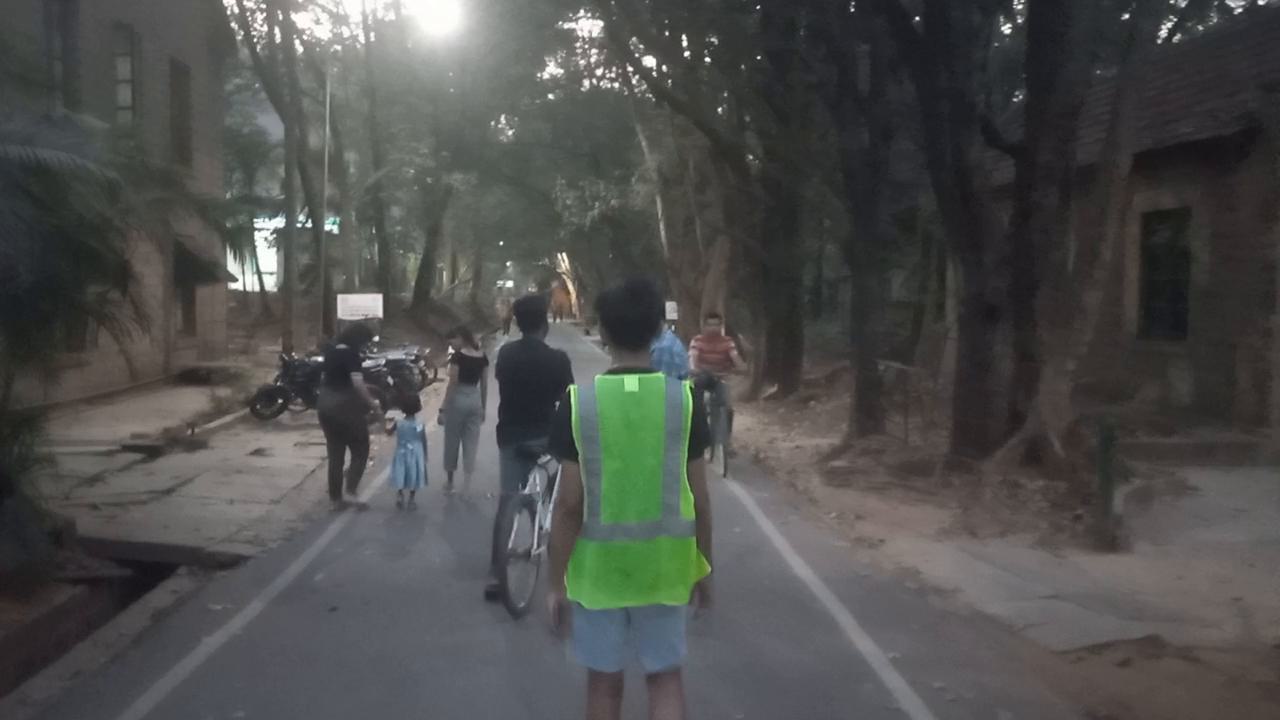}
   \label{fig:3a}
  }
  \subfloat[Object detection]{
   \includegraphics[width=0.3\columnwidth,height=2.15cm]{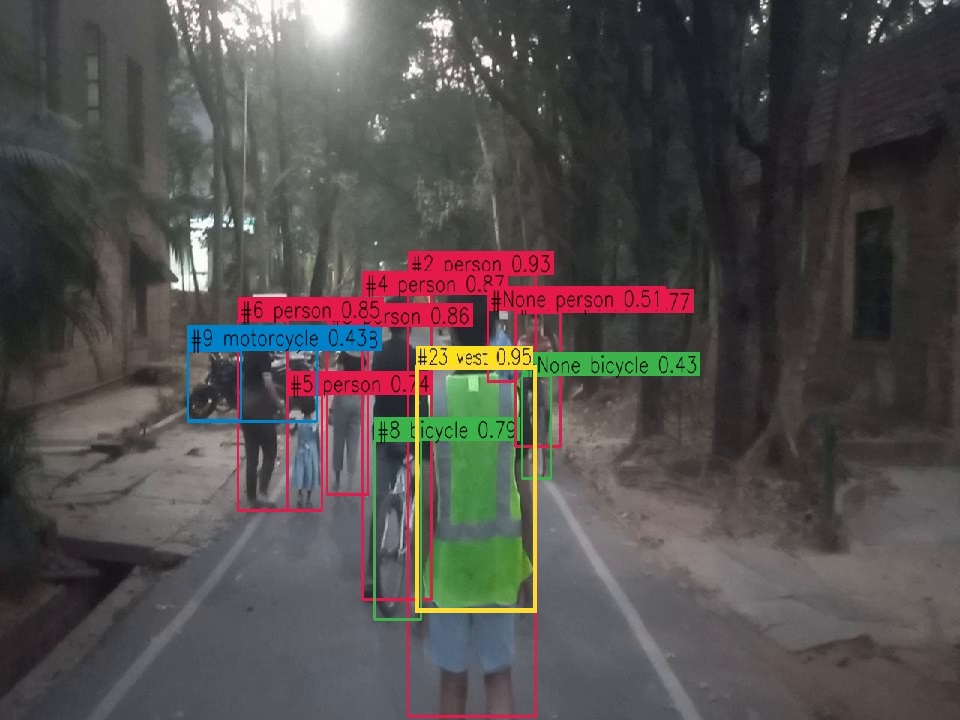}
    \label{fig:3b}
  }
    \subfloat[Road segmentation]{
   \includegraphics[width=0.3\columnwidth,height=2.15cm]{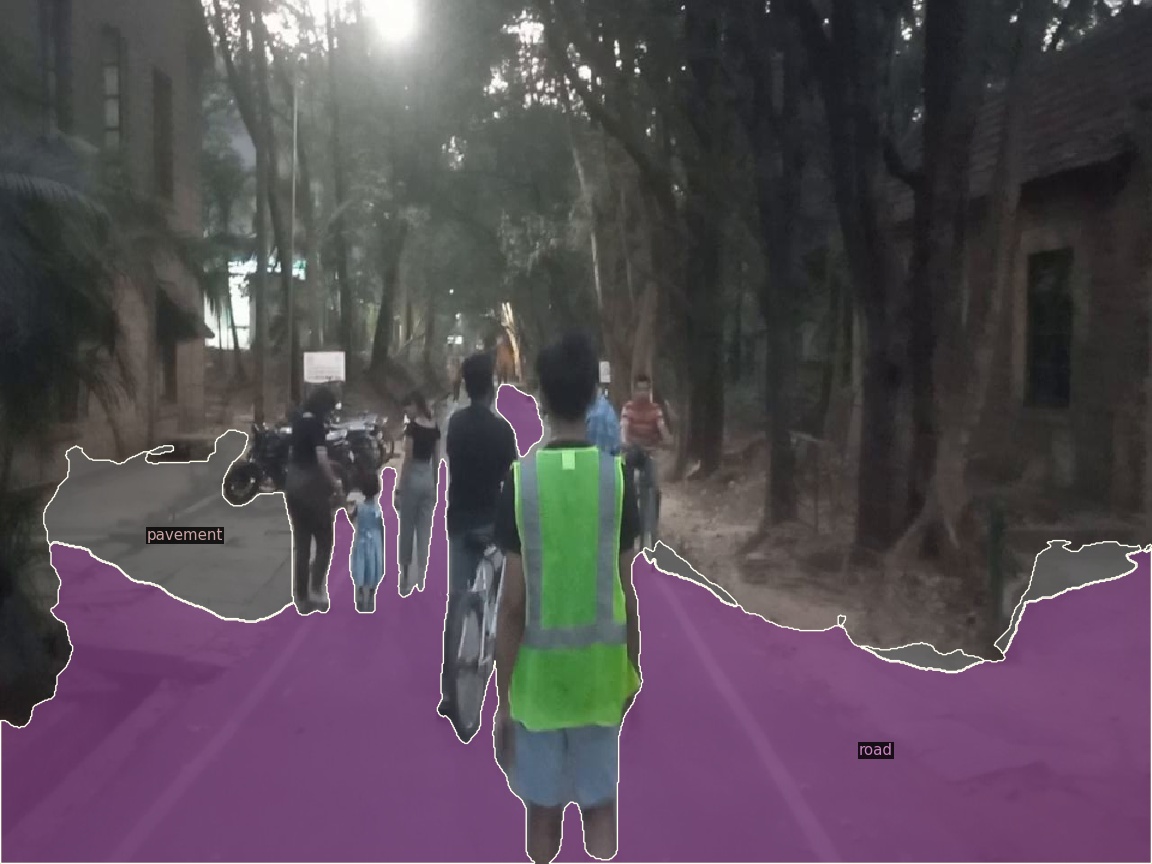}
    \label{fig:3c}
  }  \\
  \subfloat[VIP segmentation]{
   \includegraphics[width=0.3\columnwidth]{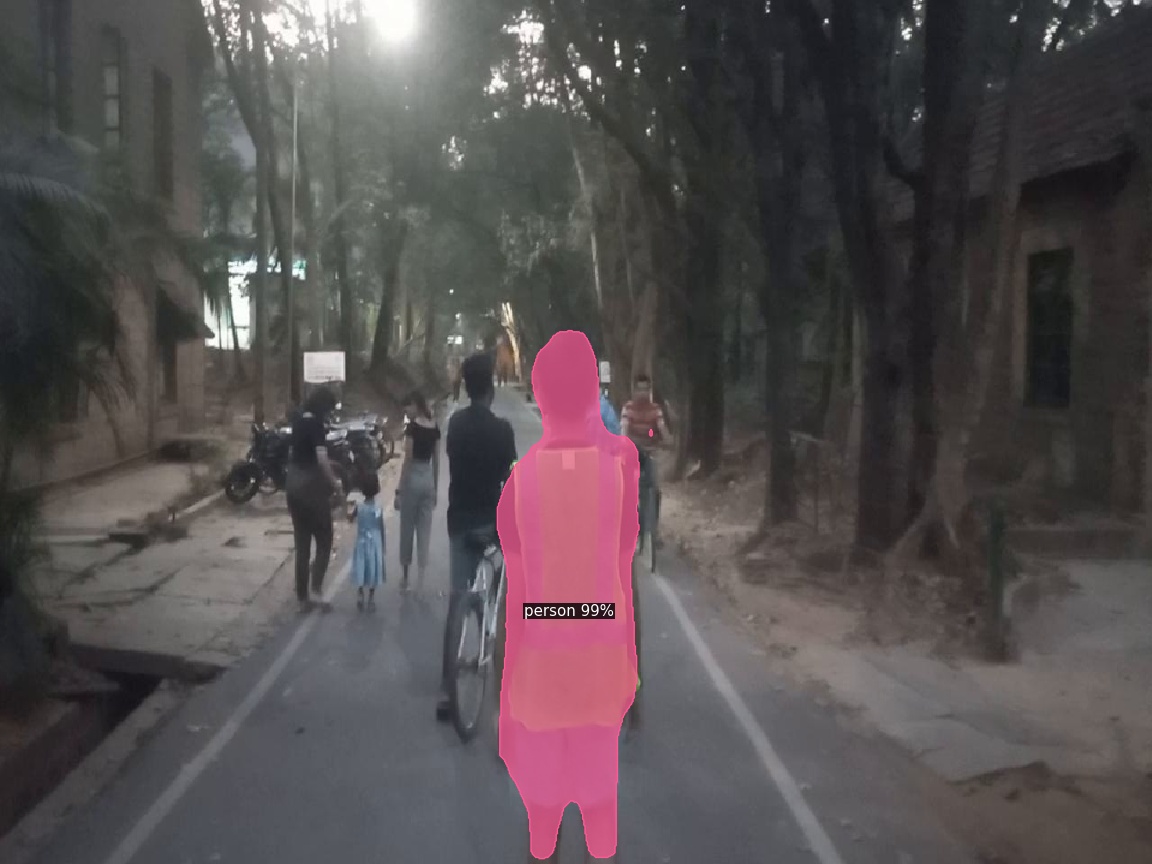}
    \label{fig:3d}
  }  
   \subfloat[Depth map]{
   \includegraphics[width=0.3\columnwidth]{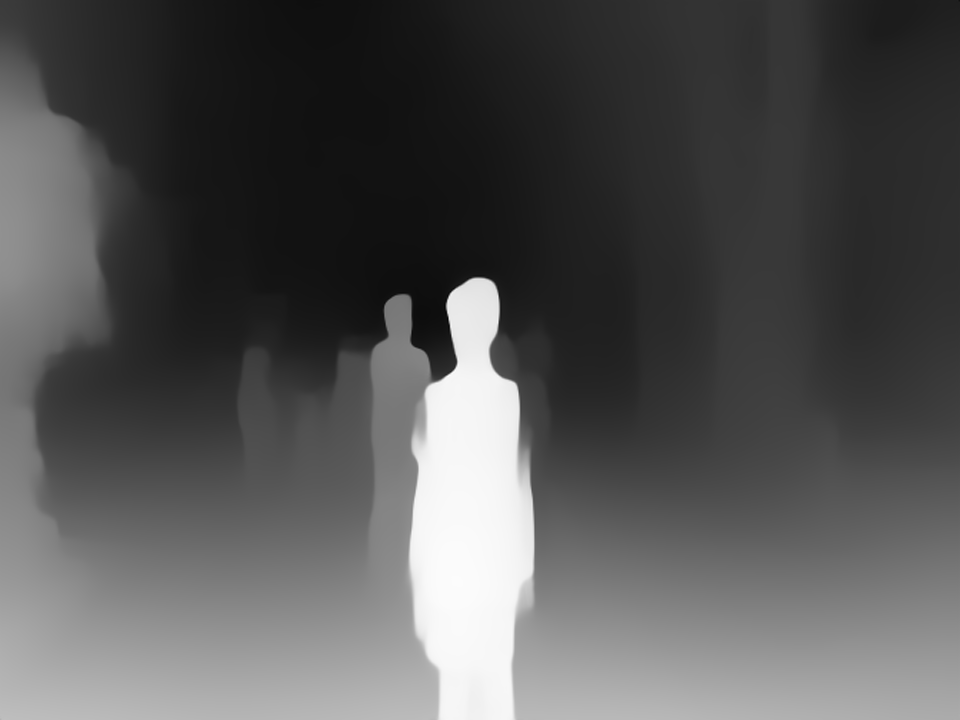}
    \label{fig:3e}
  }
  \subfloat[Required heading]{
   \includegraphics[width=0.3\columnwidth]{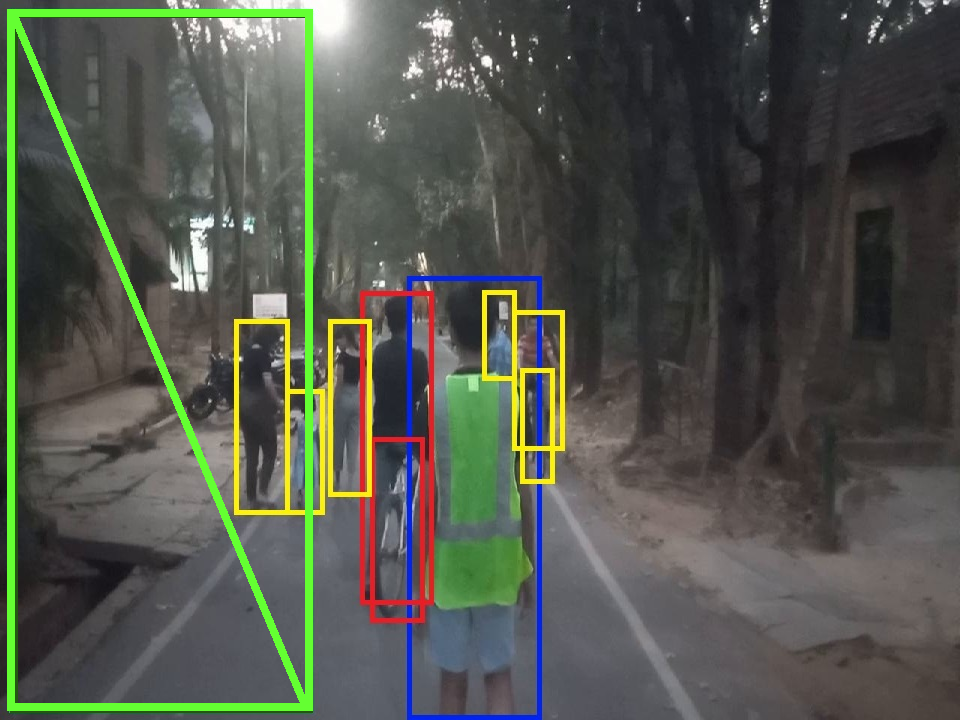}
    \label{fig:3f}
  }  
\caption{Obstacle avoidance in case of a crowded street}
\label{fig:scenario-3}
\end{figure*}

\section{Two-phase Path Planning Framework}\label{sec:framework}
Our overall planning happens as follows. The framework knows the initial (current) location of the VIP and is given the destination location to navigate to on map, in our case based on  OpenStreetMap~\cite{openstreetmaps}.
The streets in the map are represented as a weighted graph with nodes forming locations, edge serving as paths (roads, walkways), and the weight on an edge indicating the distance between the two nodes.
Given this, the global path planning uses a simple Djikstra's shortest path algorithm  to determine the proposed path between the starting and ending nodes.
In future, additional ``walkabilty'' constraints suitable for VIPs like light pedestrian traffic, fewer obstacles, fewer turns, etc. can be used used to inform this path selection.

Once the VIP sets out on this global path with the buddy drone, the local planning kicks in and guides them along this route. The UAV uses the local planning approach discussed above and performs DNN inferencing over the video feeds using accelerated edge (and optionally cloud) resources~\cite{sumanccgrid}. Small obstacles on the pavement or road such as parked cars and pedestrian traffic will suggest a slight deviation along the existing route, while a permanent obstruction, such as a road closure, causes an alternate global path to be re-planned. Here, we set the weight of the blocked road (edge) to be $\infty$ and plan a new shortest path from the current location to the destination. We illustrate this in Fig.~\ref{fig:initial-final-path} in our university campus. The initially generated path is shown in Fig.~\ref{fig:initial-path}. As the VIP proceeds (shown using markers in Fig.~\ref{fig:final-path}) a permanent obstruction is detected and an alternate global path is generated from the current location to the destination.


\section{Experiments}\label{sec:experiments}

\subsection{Implementation and Setup}
The path planning framework has been written in \textit{Python}. We use \textit{PyTorch} for invoking the various the DNN model for inferencing over the video streams as part of the local planning. We validate the effectiveness of the planning using videos captured inside our university campus using DJI Tello nano quad-copter which weighs just $80~g$ with battery. These video feeds are passed to the local planning framework to make the relevant decisions. For simplicity, we perform the inferencing on an Intel i9-10850K 20-core CPU@3.6GHz workstation with 32GB RAM, an Nvidia GeForce RTX 3060 GPU, and using CUDA v11.7. In practice, this would be replaced with a mobile edge accelerator like Nvidia Jetson AGX, Nano or Orin, coupled with cloud resources~\cite{sumanccgrid}. We use \textit{four DNN inferencing models} in the path planning framework for obstacle avoidance of the VIP and the UAV, and their performance characteristics using the workstation are shown in Table~\ref{table:dnn-config}. The inference times are reported at the $90^{th}$ percentile when running the models over 650 video frames collected at 30~fps.

\subsection{Validation}
We validate our local planning algorithm for three scenarios, each based on a $10-15$~min video stream: (1) When the VIP walks along a pedestrian footpath with tree overhands, (2) Along the side of a street with parked vehicles, and (3) On a walkway with several other pedestrians. The local planning detects the obstacle and recommends an alternate heading for the VIP instead of just walking straight ahead.

For each scenario, we show a single video frame of interest from the feed that captures the decision making logic using the different stages of the local planning. These are shown in Figs.~\ref{fig:scenario-1},~\ref{fig:scenario-2} and \ref{fig:scenario-3}. Of these, Figs.~\ref{fig:1a},~\ref{fig:2a}, and~\ref{fig:3a} are the raw video frame. The image after object detection, class labeling and bounding boxes assignment using \textit{YOLOv8} are in Figs.~\ref{fig:1b},~\ref{fig:2b} and~\ref{fig:3b}. Road masks are generated using semantic segmentation and shown in Figs.~\ref{fig:1c},~\ref{fig:2c} and~\ref{fig:3c}, while the VIP masks are generated using instance segmentation and reported in Figs.~\ref{fig:1d},~\ref{fig:2d} and~\ref{fig:3d}. The frames are converted to monochrome grayscale and a depth map is generated for the objects, displayed in Figs.~\ref{fig:1e},~\ref{fig:2e} and~\ref{fig:3e}. Finally, using the masks, detections and depth information, we estimate the distance, and based on the position of the bounding box of the VIP in the frame, we calculate the proposed alternative heading of the VIP if any obstruction was detected. The output of Algorithm~\ref{algo:obstacle-avoidance-enhanced} returns the final images, Figs.~\ref{fig:1f},~\ref{fig:2f} and~\ref{fig:3f}, which show the VIP with a blue box, \textit{dangerous} obstacles with a red box, \textit{warning} obstacles in a yellow box, and the proposed heading with a green box. 

\begin{table}
\caption{DNN Models Characteristics}
\begin{center}
\begin{tabular}{ |c | c | r | }
\hline
\textbf{Application} & \textbf{DNN Model}  & \textbf{\makecell{Inferencing\\time (in ms)}}\\ 
\hline
VIP Detection and Tracking & YOLO v8 (retrained) & 10\\ 
\hline
Object Detection and Tracking & YOLO v8  & 27\\ 
\hline
Panoptic Segmentation & Detectron2 & 108\\ 
\hline
Monocular Depth Estimation & MiDas & 16\\ 
\hline
\end{tabular}
\label{table:dnn-config}
\end{center}
\end{table}

\subsubsection{VIP walks on a footpath}
In scenario 1 (Fig.~\ref{fig:scenario-1}), the VIP encounters negligible obstructions along the footpath in the university campus in this feed. A tree projects into the pathway. While this is not a class detected by the object detection DNN model, our depth map model provides resiliency. It is able to understand that a collision is likely since the pixels corresponding to the tree lie at a proximate distance ahead and in the same partition of the image where the bounding box of the VIP is also present. Hence, the minimum pixel depth is calculated for the third partition and the heading is suggested towards the right direction (Fig.~\ref{fig:1f}). 

\subsubsection{VIP walks around parked vehicles}
Scenario 2 shown in Fig.~\ref{fig:scenario-2} has multiple parked vehicles on the side of the road. When the VIP is walking on the side of the VIP, the nearest car is shown as a dangerous obstacles (red) which is within the threshold distance of the VIP. The farther cars are shown in yellow. The obstacle avoidance algorithm generates the required heading towards the right in Fig.~\ref{fig:2f} which will avoid the parked car from the path of the VIP. 

\subsubsection{VIP walks on a crowded street}
In scenario 3 shown in Fig.~\ref{fig:scenario-3}, the object detection model detects multiple instances of the \textit{person} class. We consider other persons excluding the VIP as obstacles to VIP. Using the segmentation, we identify the VIP, and mark other person in red or yellow box depending on their proximity to the VIP. Based on the least pixel count of the first partition from the left, the VIP is required to head towards left Fig.\ref{fig:3f}.

\section{Conclusion and Future Work}\label{sec:conclusion}
In this paper, we have proposed a dual-layered path planning framework for UAVs assisting visually impaired for navigation. The local planning uses DNN inferencing on live UAV video feeds for obstacle avoidance, while ensuring that the VIP always remains in the FoV. The global plan leverages the current location of the VIP using GPS and is updated using the local plan. We validate our algorithm on various scenarios in outdoor environment and justify the feasibility of our approach. 

As a part of future work, we wish to work on more complex cases where the drone's heading can be different from the VIP's heading. We plan to extend our work to distributed path planning using video feeds captured using multiple drones. 

\bibliographystyle{IEEEtran}
\bibliography{paper}

\end{document}